%% file: main.tex
\pdfoutput=1
\documentclass[11pt]{article}

\usepackage{EACL2023}

\usepackage{times}
\usepackage{latexsym}
\usepackage{pdflscape}
\usepackage{amssymb}
\usepackage{soul}
\usepackage{float}

\usepackage[T1]{fontenc}
\usepackage[utf8]{inputenc}

\usepackage{microtype}

\usepackage{inconsolata}

\usepackage{tabularx}
\usepackage{longtable}
\usepackage{graphicx}
\usepackage{verbatim}

\usepackage{amsmath}
\DeclareMathOperator*{\argmax}{argmax}
\DeclareMathOperator*{\softmax}{softmax}

\title{Disentangling continuous and discrete linguistic signals \\ in transformer-based sentence embeddings}

\author{ Vivi Nastase \and Paola Merlo
\vspace{1.5mm} \\
Department of Linguistics \\ University of Geneva \\ \texttt{Paola.Merlo@unige.ch, vivi.a.nastase@gmail.com} 
}

\begin{document}
\maketitle
\begin{abstract}
Sentence and word embeddings encode structural and semantic information in a distributed manner. Part of the information encoded -- particularly lexical information -- can be seen as continuous, whereas other -- like structural information -- is most often discrete. We explore whether we can compress transformer-based sentence embeddings into a representation that separates different linguistic signals -- in particular, information relevant to subject-verb agreement and verb alternations. We show that by compressing an input sequence that shares a targeted phenomenon into the latent layer of a variational autoencoder-like system, the targeted linguistic information becomes more explicit. A latent layer with both discrete and continuous components captures better the targeted phenomena than a latent layer with only discrete or only continuous components. These experiments are a step towards separating linguistic signals from distributed text embeddings and linking them to more symbolic representations.
\end{abstract}

\section{Introduction}

As deep learning models become more and more powerful, the need grows to move away from black box models to interpretable ones. An important reason for this is that black box models may make good predictions for the wrong reasons. There is a big risk involved with deploying such models in environments where wrong predictions can have dire consequences \cite{Rudin2021InterpretableML}. 

Explanations need to be formulated based on the conscious primitives of language. The expressive power of human thought and language are arguably built by compositional processes that operate on objects that, at least at the conscious level, are symbolic. 

At a high level, the discrete, symbolic and combinatorial nature of language needs to be reconciled with the statistical patterns and the machine encoding of language in distributed representations. 

At a lower level, understanding the representations of words, sentences, and text  produced with deep learning models would help trace the different syntactic and semantic signals and explain how they are encoded in distributed representations. 

Information in the input word or text fragment is encoded into a vector of fixed dimensions with continuous values. Some of the information encoded can be viewed as continuous. For example, our intuitive understanding of lexical semantic  properties is conceived as a similarity space so that we can judge whether words or text fragments are close or distant. Other types of information -- e.g. grammatical number, gender, roles, verb classes -- are more discrete in nature. While the good results in using these representations for various NLP tasks \cite{wang-ea2019,rajpurkar-etal-2018-know} indicate that both discrete and continuous information is encoded in these representations, it is not explicit.

Unlike previous work, we do not aim to show that sentence embeddings encode information pertaining to specific linguistic phenomena, but to detect how such information is encoded, and whether we can disentangle different linguistic signals from transformer-based sentence embeddings. Because sentence representations compress a multitude of linguistic information, we use datasets that focus on and encode specific linguistic phenomena  -- in particular, subject-verb agreement and verb alternations -- as commonly done \cite{nikolaev-pado-2023-representation,linzen2016}. To test how well we can detect signals relevant to these (implicitly) provided phenomena, we use a variational autoencoder-based system. In previous work we have established that by reshaping the sentence embeddings induced using Transformer models from a 1D array to a 2D array has a positive impact on detecting patterns that signal subject-verb agreement, and we explored different system variations \cite{nastase-merlo2023}. In this work we show that a latent layer that has a continuous and a discrete part leads to best results. By analysing the kind of errors the system makes when masking different parts of the latent layer, we show that they encode different types of information. The code and data are available at \url{https://github.com/CLCL-Geneva/BLM-SNFDisentangling}.

%We test RoBERTa \cite{liu2019roberta} and Electra \cite{clark2020electra} sentence embeddings, combined with baseline FFNN and VAE-based architectures, and show that having a joint latent layer -- with a discrete and a continuous part -- in a VAE-based system allows for better modeling of the targeted grammatical phenomena compared to using only a continuous or only a discrete latent vector.

%\paragraph{Terminology}
%\textcolor{blue}{(This is the same as Nastase and Merlo shouldn't we cite?)}
%Sentence embeddings can be read from the output of BERT systems as a $1\times N$  vector (N usually 768 or 1024). This can be viewed as the projection of the sentence into an N-dimensional space. In this paper, we use the word {\it dimensions} to refer to the shape of the data structure used to represent the sentence embeddings. In particular, we use {\it one-dimensional array} to refer to the $1\times N$ vector sentence representation obtained directly from BERT, and {\it 2D representations} to refer to 2D reshaped ($Rows \times Columns$) vector. 

\section{Related work}

%%VV: from Flehi, 2023, Interpretable Sentence Representation with Variational Autoencoders and Attention
%%paraphrase more

Neural representations have lead to breakthroughs in multiple tasks, including NLP, but they, and the models used to build them, are quite opaque. Neural systems may produce the correct answer but for the wrong reason, or based on spurious correlations in the input. Understanding the neural network blackboxes and the representations they induce or learn is a crucial research direction \cite{bengio-ea2013}. \newcite{Rudin2021InterpretableML} provide an overview of interpretable ML, which include disentanglement techniques. Disentanglement can also be used to design and select input data such that it covers the targeted interpretable concepts and help improve generalization \cite{pmlr-v119-locatello20a}.

Disentanglement, often implemented using Generative Adversarial Networks (GANs) \cite{goodfellowNIPS2014} and Variational AutoEncoders (VAEs) \cite{schmidhuber1992,kingma2013vae}, has found several applications in NLP, as it can help separate the various types of information encoded in a sentence, such as syntax and semantics \cite{chen-etal-2019-multi,bao-etal-2019-generating}, text style \cite{Fu2018style-transfer,john-etal-2019-disentangled} or morphological information \cite{zhou-neubig-2017-multi}. The representation on the latent layer can have continuous or discrete variables. Continuous representations can also be disentangled \cite{higgins2017beta,mathieu2019disentangling,chen2615isolating}, while the discrete one by default separates specific factors. 

\citet{bao-etal-2019-generating} and \citet{chen-etal-2019-multi} use two continuous variables to model semantic and syntactic information on the latent layer of a VAE. \citet{bao-etal-2019-generating} enforce the encoding of syntactic information in the latent layer by predicting the linearized parse tree of the input. \citet{chen-etal-2019-multi} use multi-task training to encourage the separation of information on the latent layer.

%\cite{mitchell-2016-decomposing,mitchell-steedman-2015-orthogonality}, style \cite{tenenbaum1996}, or even the compositionality function \cite{socher-etal-2012-semantic}. 

%Disentanglement can be supervised, when the targeted underlying factors are explicitly given, or unsupervised when for various reason, the underlying factors cannot be specified. For such settings,  are often used. 
%They try to learn a mapping from points in one probability distribution to points in another. The first distribution is over points in the latent space. The second distribution is over the space from which the data are drawn (e.g., random natural images in the space of natural images). The statistical independence between the latent features makes it easy to disentangle the representation \cite{bengio-ea2013}: a disentangled representation simply guarantees that knowing or changing one latent feature (and its corresponding concept) does not affect the distribution of any other. 

\newcite{mercatali-freitas-2021-disentangling-generative} learn to isolate 9 generative factors using a variational autoencoder (VAE) architecture with Gumbel-softmax sampling \cite{jang2017categorical}. Sentences are encoded (and decoded) using an LSTM.
\newcite{zheng-lapata-2022-disentangled} propose a different method for disentangling relations expressed in a sentence which may share arguments. This is implemented as an extension to sequence-to-sequence (seq2seq) models, where at each decoding step the source input is re-encoded by conditioning the source representations on the newly decoded target context. These specialized representations make it easier for the encoder to exploit relevant-only information for each prediction.

\citet{huang-etal-2021-disentangling} disentangle syntactic and semantic representation using a sentence encoder and a parse encoder. Learning to produce paraphrases of the input sentence with the given parse structure forces the sentence encoder to produce a semantic representation devoid of syntactic information.

We build on \citet{dupont2018learning}, who shows that a combination of discrete and continuous factors characterizing images can be learned in an unsupervised manner. We experiment with different representations on the latent layer of a VAE-like system, to test whether specific grammatical information can be disentangled from transformer-based sentence embeddings.

%% add "Frequency effects on syntactic rule learning in transformers" (Wei et al. 2021)?

\section{Grammatical phenomena to study sentence representations}

We investigate whether specific grammatical information can be accessed from distributed sentence representations. 
Sentences are combinations of linguistic phenomena, which LLMs compress in fixed-length continuous vectors. Because of this, linguistic phenomena are often studied on specifically designed or selected datasets (e.g. \cite{nikolaev-pado-2023-representation,linzen2016}), that isolate or emphasize the targeted phenomena.% while reducing the degrees of freedom of other linguistic variables. 
We also use artificially generated datasets, Blackbird Language Matrices (BLMs) \citep{merlo-ea2022arxiv,merlo2023}, inspired by Raven Progressive Matrices visual pattern tests that rely on the solver detecting overlapping rules \cite{raven1938,chi2019RAVEN}.

\subsection{Input data}
\label{sec:data}

A Blackbird Language Matrix (BLM) problem \citep{merlo2023} has an input consisting of a context of $S$ sentences that share the targeted grammatical phenomenon, but differ in other aspects relevant for the phenomenon in question. 
BLMs are multiple-choice problems, and each input is paired with a set of candidate answers, where the incorrect ones are built by corrupting some of the generating rules of the input sequence. This added dimension of the datasets facilitates the investigation of the kind of information a system is able to disentangle from the sentence embeddings.

BLM datasets can also vary in lexical complexity. The datasets usually comprise three levels of complexity. Type I data is generated based on manually provided seeds, and a template for its generative rules. Type II data is generated based on Type I data, by introducing lexical variation using a transformer, by generating alternatives for masked nouns. Type III data is generated by combining sentences from different instances from the Type II data. This allows investigations into the impact of lexical variation on the ability of a system to detect grammatical patterns.

We use two BLM datasets, which encode two different linguistic phenomena, each in a different language: subject verb agreement in French, and an instance of verb alternations in English.

\paragraph{BLMs for subject-verb agreement in French}

Subject-verb agreement is often used to test the syntactic abilities of deep neural networks \cite{linzen2016,gulordava2018,goldberg2019,linzen2021}. While
theoretically simple, it can have several complicating factors: e.g. linear or structural distance between the subject and the verb.%, intervening nouns and verbs.% called attractors.

\input{BLM-example}

We use BLM-AgrF \cite{an-etal-2023-blm},\footnote{The data is publicly available at \url{https://github.com/CLCL-Geneva/BLM-SNFDisentangling}} illustrated in Figure \ref{BLM-agreement}. The input for each instance consists of a context set of seven sentences that share the subject-verb agreement phenomenon, but differ in other aspects -- e.g. number of intervening attractors between the subject and the verb, different grammatical numbers for these attractors, and different clause structures.

\paragraph{BLMs for verb alternations in English}

The study of the argument-structure properties of verbs and semantic role assignments is also a test-bed for the core syntactic and semantic abilities of neural networks \citep{kann-etal-2019-verb,yi-etal-2022-probing}. In particular, \citet{yi-etal-2022-probing} demonstrates that transformers can encode information on the two alternants of the well-studied \textit{spray-load} alternation \citep{rappaport1988theta,levin1993}. We use the dataset BLM-s/lE \citep{samo-etal-2023}, whose structure is exemplified in Figure \ref{BLM-sprayload}.

\input{BLMsl-example}

As can be seen, a BLM instance consists of a context set comprising one alternant of the \textit{spray-load} alternation and other sentences that provide the syntactic properties of the arguments (e.g. passivization strategies). The target sentence is the other alternant (whose arguments share common properties with the first sentence) to be chosen from an answer set of superficially minimally, but, syntactically and semantically deeply, different candidates. (See \citet{samo-etal-2023} for more detail.)

There are two groups within this dataset, one for each of the two alternates. {\it Group 1} (ALT-ATL) has the alternant \textsc{Agent}-\textsc{Locative}-\textsc{Theme} (e.g. \textit{The girl sprayed the wall with the paint) } in the context and the correct answer is the alternant whose configuration is \textsc{Agent}-\textsc{Theme}-\textsc{Locative} (e.g. \textit{The girl sprayed paint onto the wall}), while the the template of {\it Group2} (ATL-ALT) starts with \textsc{Agent}-\textsc{Theme}-\textsc{Locative} and the target answer is \textsc{Agent}-\textsc{Locative}-\textsc{Theme}. %Group 1 data is the data produced from the matrix in Figure \ref{BLM-sprayload}. 

\paragraph{Datasets statistics}
Table \ref{tab:data} shows the datasets statistics. Each subset is split 90:10 into train:test subsets. 20\% of the train data is used for development.

\begin{table}[h]
    \begin{tabular}{l|r|rr} \hline
     & Subj.-verb agr. & \multicolumn{2}{c}{Verb alternations} \\ 
     &                 & ALT-ATL & ATL-ALT \\ \hline
    Type I  & 2304  &  3750 & 3750 \\
    Type II & 38400 &  15000 & 15000 \\
    Type III & 38400 &  15000 & 15000 \\ \hline
    \end{tabular}
    \caption{Types I, II, III correspond to different amounts of lexical variation within a problem instance.}
    \label{tab:data}
\end{table}

\subsection{Sentence representations}
\label{sec:sentence}

We investigate sentence embeddings obtained from two transformer-based systems:  RoBERTa \cite{liu2019roberta} and Electra \cite{clark2020electra}, with a FFNN baseline and an encoder-decoder architecture inspired by variational autoencoders, represented schematically below.

%\vspace{0.5cm}
    \includegraphics[width=0.45\textwidth]{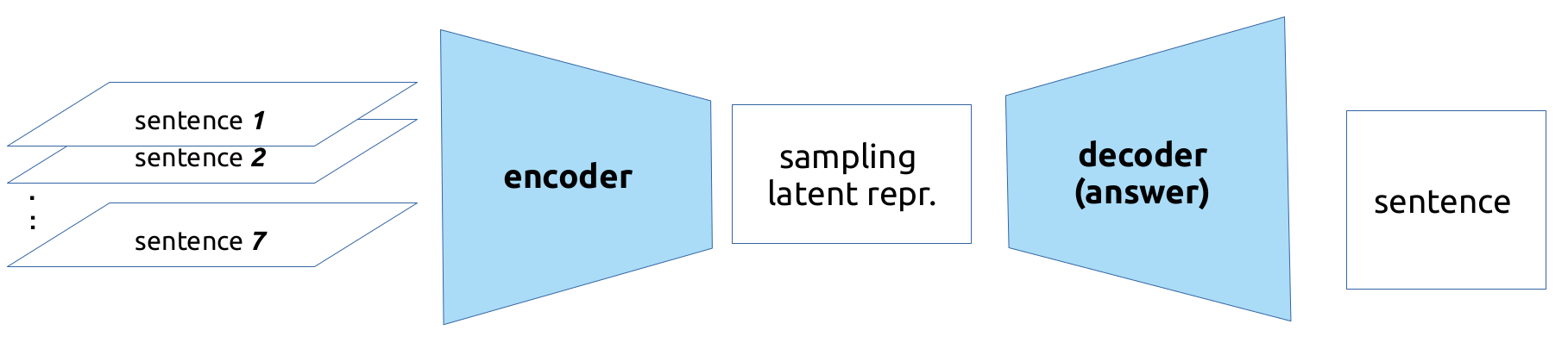} 
%\vspace{0.5cm}

For all of these systems we use as sentence embedding the encoding of the \textit{[CLS]} or the \textit{<s>} character read from the last layer of the model. 

\subsection{Detecting linguistic signals in sentence embeddings}

We explore sentence embeddings using a baseline FFNN and variations of a system based on the variational autoencoder architecture.\footnote{The code will be made publicly available upon publication.}
%, publicly available on github\footnote{ \url{https://github.com/CLCL-Geneva/BLM-SNFDisentangling}}. 
The system's hyperparameters -- parameters of the CNN and FFNN layers in the encoder and decoder -- were established using development data on the subject-verb agreement problem, using type I data for training and testing. It was then deployed on the other train/test configurations and the verb alternation problem. We add to the encoder-decoder architecture different sampling methods on the latent layer of the encoder-decoder --  continuous, discrete and joint sampling -- to test whether separating discrete and continuous components makes the targeted phenomena more explicit.

\subsubsection{FFNN baseline}
The FFNN baseline is a three-layer feed-forward neural network, that maps an input sequence of sentence embeddings into a vector representing the answer. The learning objective is to maximize the probability of the correct answer from the candidate answer set and is implemented through the max-margin loss function. This function combines the scores of the correct and erroneous sentences in the answer set relative to the sentence embedding predicted by the system: \\

\noindent $ loss_a(x) =$ 

$ \sum_{e_i} [1 - score(e_c, e_{pred}) + score(e_i,e_{pred})]^{+}$ \\

\noindent where $e_i$ is the embedding of a sentence $a_i$ in the answer set $\mathcal{A}$, $e_{pred}$ is the embedding produced by the system for input $x$, and $score$ is the cosine of the angle between the given vectors.

For prediction, the answer $a_i$ with the highest $score$ from a candidate set w.r.t. the produced sentence embedding is taken as the correct answer.

\subsubsection{Encoder-decoder}

This system is similar to a variational autoencoder (VAE) \cite{kingma2013vae,kingma2015}, but the decoder does not reconstruct the input, rather it constructs an answer. 

A variational autoencoder encodes an input sequence into a compressed representation, and then attempts to reconstruct it, while modeling the compressed representation of the input as a distribution over the latent space, rather than a single point. This procedure avoids overfitting and ensures that the latent space is structured and thus has good properties that enable the generative process.

%\vspace{0.5cm}
%\includegraphics[width=0.45\textwidth]{figs/VAE.pdf}
%\vspace{0.5cm}

The input is a stack of 2D-ed sentence embeddings. The encoder consists of a 3D CNN layer with a 3x15x15 kernel for the input SxNxM where S is the length of the input sequence (7) and NxM is the shape of the 2D sentence representation array (we use 32x24). This is followed by a linear layer that compresses the output of the CNN to the dimension set for the latent layer. The decoder consists of a linear layer followed by a CNN (with a 1x15x15 kernel) that produces a 2D array representing the embedding of the predicted answer.

The objective function of the VAE captures the modeling (reconstruction of the input) and regularization constraints placed on the latent space through two factors:  
~\\ ~\\
\noindent {\small $
\mathcal{L}(x)=\mathbb{E}_{q_\Phi(z|x)}[log~p_\Theta(x|z)]-KL(q_\Phi(z|x)\|p(z)) $} \\

This is implemented through the corresponding loss function, where $x$ is the input and $x'$ is output, i.e. the reconstructed input.
~\\ ~\\
\noindent $loss(x) = \|x - x'\|^2 + KL(q_\Phi(z|x)\|p(z))$ \\

Because our system does not reconstruct the input but rather outputs a sentence embedding, the loss function becomes:
~\\ ~\\
\noindent $loss(x) = loss_a(x) + KL(q_\Phi(z|x)\|p(z))$ \\

\noindent where $loss_a$ is the max-margin loss function used by the baseline FFNN.

We can enforce different assumptions on the latent layer, and sample a vector accordingly from the output of the encoder. In particular, we can consider the latent layer to be a continuous vector, a discrete one, or a combination. For each variation the KL divergence factor will change accordingly, and $loss_a(x)$ will remain the same.

\paragraph{Continuous}

In this setting, the assumption is that the vector on the latent layer is a vector of continuous values, with a standard Gaussian prior distribution $p(z) = \mathcal{N}(0,1)$. The output of the encoder is a vector interpreted as $[\mu_x; \sigma_x]$ modeling a normal distribution from which the vector $z$ is sampled: $z \sim q_\Phi(z|x) = \mathcal{N}(\mu_x, \sigma_x)$ \cite{kingma2013vae}. The KL factor becomes $KL(\mathcal{N}(\mu_x, \sigma_x)\|\mathcal{N}(0,1))$.

%To be able to implement this operation in a deep learning framework -- sampling introducing discontinuity in the process -- \cite{kingma2015} have proposed the reparametrization trick, where $z = \mu_x + \sigma_x \cdot \epsilon$, and $\epsilon \sim \mathcal{N}(0,1)$ and backpropagation operates normally over $[\mu_x; \sigma_x]$, the output of the encoder.

\paragraph{Discrete}

To model data that may have discrete structure, \newcite{jang2017categorical} introduce the Gumbel-Softmax distribution, which can approximate categorical samples. If $c$ is a categorical variable with class probabilities $\pi_1, ... \pi_k$, drawing a sample $c$ from a categorical distribution with class probabilities $\pi$ would be:
~\\ ~\\
\noindent $c \sim one\_hot(\argmax_i[g_i + log \pi_i]) $
~\\

\noindent where $g_i \sim Gumbel(0,1)$, and the nondifferentiable $argmax$ funtion is approximated using $softmax$:
\[
\begin{split}
\argmax_i[g_i + log \pi_i] \approx \softmax_i[g_i + log \pi_i] =\\
= \frac{exp((g_i + log \pi_i)/\tau)}{\sum_{j=1}^k exp((g_j + log \pi_j)/\tau)}
\end{split}
\]

\noindent where $\tau$ is a $temperature$ that controls the softmax distribution: higher values result in more uniform distributions, whereas for values closer to 0 the expected value approaches the expected value of a categorical random variable with the same logits.
The KL factor becomes $KL(q_\Phi(c|x)\|Gumbel(0,1)) $

\paragraph{Joint}

A latent vector with a discrete and continuous part can also be used \cite{dupont2018learning}. In this case the encoder models a distribution with continuous latent $z$ and discrete latent $c$ as $q_\Phi(z,c|x)$ with prior $p(z,c)$ and likelihood $p_\Theta(x|z, c)$. Because the continuous and discrete channels can be assumed to be conditionally independent, $q_\Phi(z, c|x)  = q_\Phi(z|x) q_\Phi(c|x)$; $p(z,c) = p(z) p(c)$ and $p_\Theta(x|z,c) = p_\Theta(x|z) p_\Theta(x|c)$, where each of the probabilities and samplings will be done according to the continuous or the discrete sampling respectively. The KL factor becomes \\
~\\
$KL(q_\Phi(z,c|x)\|p(z,c)) = \\
 KL(q_\Phi(z|x)\|p(z)) + KL(q_\Phi(c|x)\|p(c))$.

\section{Experiments}

We hypothesize that we can separate different types of linguistic information, specifically lexical from structural information, in transformer-based sentence representations. We test this hypothesis through two types of analysis.

\begin{description}
    \item[A1] Through the performance on the BLM multiple-choice problems that encode different linguistic phenomena, in two different languages.
    \item[A2] Through error analysis, which will reveal:
    \begin{description}
        \item[A2.1] what kind of information is accessed in sentence embeddings to solve the problems;
        \item[A2.2] whether different types of information is captured in the discrete and continuous parts of the latent layer.
    \end{description}
\end{description}

Should our hypothesis be correct, we expect analysis {\bf A1} to show higher performance for joint sampling on the latent layer of our encoder-decoder system, compared to either discrete or continuous sampling alone. The different range of lexical variation of the three dataset subsets (type I, II, III) adds another dimension to the investigation: lexical variation, a source of continuous information in neural networks, allows us to assess the impact of such information on the differentiation of the input into discrete and continuous signals. We measure performance in terms of F1 score, and report averages over 5 runs.

Analyses {\bf A2.1} and {\bf A2.2} investigate the kind of errors the system makes when using different variations of the system. The erroneous candidate answers represent different types of errors -- structural or lexical -- and changes in the frequency of such types of errors provide additional clues regarding the information encoded in the different parts of the latent vector.

\paragraph{Data} We use the data described in Section \ref{sec:data}, and sentence embeddings generated using RoBERTa \cite{liu2019roberta} and Electra \cite{clark2020electra} pretrained models\footnote{RoBERTa: \textit{xlm-roberta-base}, Electra: \textit{google/electra-base-discriminator}}. 

For space reasons, we show here results when training on type I, II and III -- increasing lexical variation -- and test on type III -- maximal lexical variation. This is the most difficult learning set-up, and will allow us to test whether the system can discover robust patterns, or rather it picks up on lexical regularities.

\paragraph{System}
We analyze the effects of compressing these embeddings into low-dimensional representations, with discrete and continuous components, using the system described in Section \ref{sec:sentence}. Unlike previous work on disentangling syntax and semantics \cite{chen-etal-2019-multi,bao-etal-2019-generating,huang-etal-2021-disentangling}, the targeted grammatical information is only implicitly provided.

%\textcolor{blue}{ \textcolor{red}{Why anonymous?}
%\pgm{I don't think it should be anonymous, it is actually more of a give away than not. It is published and easily acccessed.  Also, I think this aspect only deserves a reference. }

Previous work \cite{nastase-merlo2023} explored how the subject-verb agreement information can be accessed in BERT \cite{devlin-etal-2019-bert} sentence embeddings. Experiments with various architectures have shown that this information seems to be regularly distributed in the sentence embedding (the embedding of the [CLS] special token), such that reshaping the one-dimensional array corresponding to the sentence embedding into a 2D-array makes the grammatical pattern more easily accessible.

%%VV explain here about not tuning the system for each dataset and set-up -- we expect linguistic information to be regularly distributed (and that seems to be confirmed by the experiments)

%\pgm{I think this way of presenting the current work is too reductionist. It is more original than you say.}

We adopt a similar experimental set-up, using a VAE-based architecture. Our goal is to determine whether we can separate different types of linguistic information in sentence embeddings, in a general framework. For this reason we use the same architecture for both datasets, and do not tune hyperparameters or adjust the system set-up for each dataset, or for different sentence embeddings. We use the hyperparameters tuned using BERT \cite{devlin-etal-2019-bert} sentence embeddings, with sentence embeddings reshaped as 32x24 arrays, trained and tested with type I subject-verb agreement data. The size of the latent layer for continuous sampling is 5. For the joint sampling we use 1x2+5 (7) and 2x2+5 (9) sized vectors (1 and 2 binary categories, continuous portion length 5). We include experiments using a continuous latent that matches the length of the vector on the latent layer for the joint sampling (7 and 9) to show that the increase in performance is not due to a longer vector on the latent layer.

All systems used a learning rate of 0.001 and Adam optimizer, and batch size 100. The training was done for 120 epochs. The experiments were run on an HP PAIR Workstation Z4 G4 MT, with an Intel Xeon W-2255 processor, 64G RAM, and a MSI GeForce RTX 3090 VENTUS 3X OC 24G GDDR6X GPU.

\paragraph{A1: Performance analysis}

We analyze first the performance of different variations of the encoder-decoder system in terms of F1 score averages over 5 runs\footnote{The standard deviation for all set-ups is lower than 1e-03, so we do not include it.}. Should having a discrete part of the latent layer be useful, we expect the set-up using joint sampling to have the highest performance. 
To control for alternative explanations of the improvement, we set up pairwise comparisons with two control models: we compare joint models to simple models with latent vectors of equal size, and with the simple model that has overall highest performance. We verify the robustness of the results across two word embedding representations. 

\begin{figure}[h!]
    \centering
    {Subject-verb agreement}
    \includegraphics[trim={0 1cm 0 0},clip,width=0.49\textwidth]{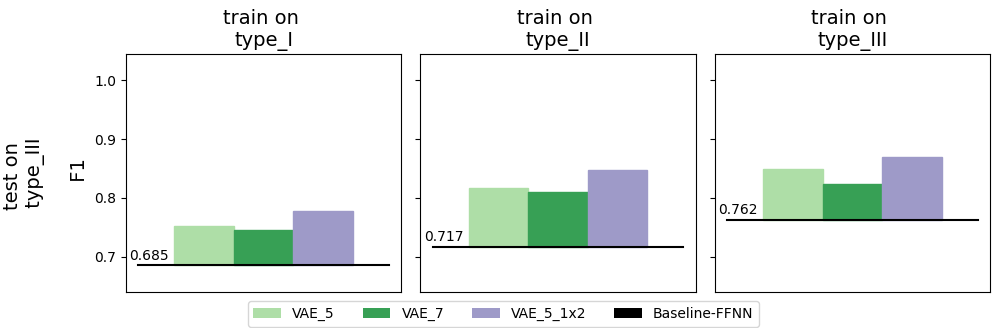}
    {Verb alternations ALT-ATL}
    \includegraphics[trim={0 0 0 0},clip,width=0.49\textwidth]{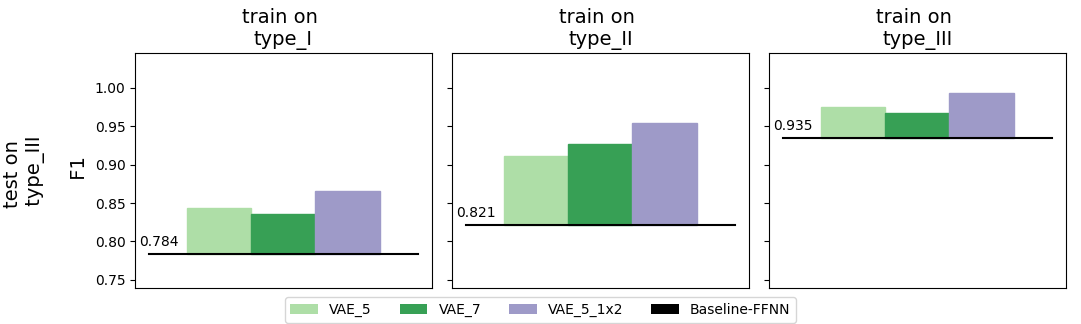}    
    \caption{F1 (avg. over 5 runs): continuous and joint sampling, Electra sentence embeddings.}
    \label{fig:agr_sampling}
\end{figure}

The comparative results of selected set-ups are shown in Figure \ref{fig:agr_sampling}. They cover: the FFNN baseline; VAE\_5, the system with continuous sampling and latent size 5 (the best variation when using continuous sampling); VAE\_7, the system that has the latent layer size equal to the one used in joint sampling; VAE\_5\_1x2, the system using joint sampling, with a continuous part of length 5, and 1 binary category for capturing discrete signals. All results were obtained for Electra sentence embeddings. The results on all configurations and both sentence embedding types are shown in Figure \ref{fig:agr_sampling-g2} in the Appendix.

\begin{figure*}[h!]
    \centering
    {Subject-verb agreement}
    \includegraphics[trim={0 1.5cm 0 0},clip,width=0.98\textwidth, height=4cm]{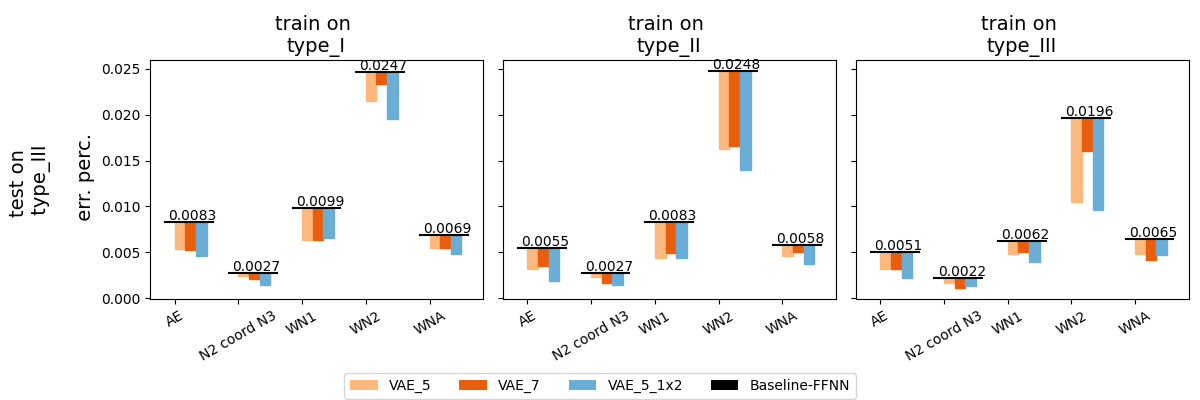}
    {Verb alternations ATL-ALT}    
    \includegraphics[width=0.98\textwidth,height=4cm]{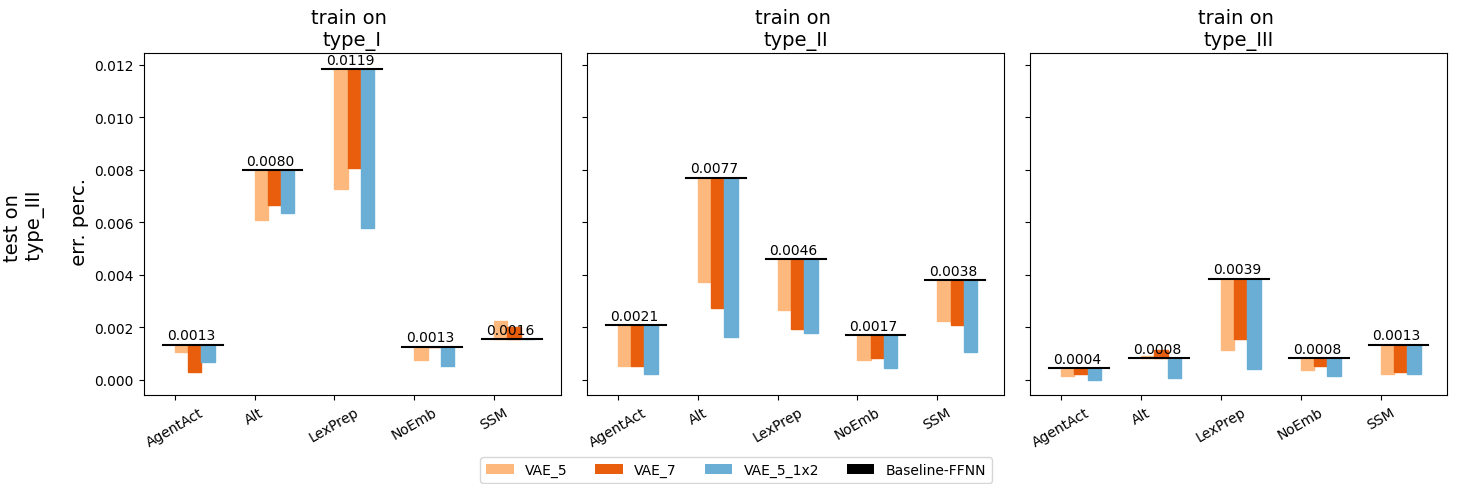}
    \caption{Error analysis for continuous and joint sampling using Electra sentence embeddings.}
    \label{fig:errors}
\end{figure*}

Joint sampling leads to highest performance on all datasets and sentence embeddings, and particularly in the more difficult set-up of using maximal lexical variation data (type III), as expected. 
Using a bigger continuous latent layer leads to lower performance, showing that the increase in performance when using joint sampling is indeed due to having a discrete portion. This indicates that this configurations captures more, or more explicitly, linguistic information that is relevant to the two phenomena represented in the datasets.

\paragraph{A2.1: Error analysis}

To understand better the kind of information the system accessed in the sentence representations, we use the answer sets, which have been constructed to include specifically built erroneous answers. Using two different problems with different properties allows us some interesting controlled pairwise comparisons. In the agreement problems, basically all incorrect answers violate structural rules. In the verb alternation problem, NoEmb and LexPrep are lexical rules, while the others are structural.

%The error analysis is presented in Figure \ref{fig:errors}, for sentence embeddings obtained using Electra, which had overall better performance\footnote{Average 0.95 vs. 0.91 for verb alternations, 0.866 vs 0.871 for subject-verb agreement.}. The pattern of errors is the same between simple and joint sampling, for both datasets. However, the joint is numerically always a little better than the simple, for all mistakes for all data.

Results are shown in Figure  \ref{fig:errors}, for sentence embeddings obtained using Electra, which had overall better performance\footnote{Average 0.95 vs. 0.91 for verb alternations, 0.866 vs 0.871 for subject-verb agreement.}.
For the agreement data, the main sources of error are WNA, WN1 and WN2. These mistakes indicate a lack of understanding of the structural aspect of agreement, preferring a linear interpretation. These are mistakes that show that the global pattern of agreement over the whole BLM, which is purely formal, has not been learnt. For all, the highest drop (compare the red and the blue bars) is obtained for the configuration that includes a discrete part in the latent layer, and most obviously for the WN2 error -- the closest NP carries the number that allows it to agree with the verb-- which humans also make. This indicates that using joint sampling allows the system to find longer distance patterns, and not be tricked by proximity.

Most of the errors specific to Alternations are in the syntax-semantic mapping (SSM), for both groups. Group 2 also shows some structural mistakes if not enough lexical variation is seen in training. This pattern of mistakes suggests that the syntax-semantic mapping, the core of argument structure, has not been fully mastered. 
When using joint sampling, the most affected mistakes (as shown in comparing the blue and red bars in Type I and type III, which reveal the clearest patterns) are the lexical ones, LexPrep, as expected.

%se type of errors drop more than when using simple sampling, indicating that the discrete part captures some linguistically relevant information for this phenomenon. Specifically

\paragraph{A2.2: Discrete vs. continuous analysis} To get closer to understanding what kind of information is encoded in the discrete and continuous portions of the latent layer, we mask these one by one (set it to 0), and perform error analysis. To analyze the change in error patterns we compare the system predictions before and after masking through Cohen's $\kappa$. Pairwise agreement between the normal system setting, and masked discrete, and 1-5 continuous layer units are presented in Figure \ref{fig:masked_agreement}. Absolute error plots are shown in Figure \ref{fig:masked error} in the appendix.

\begin{figure}[h!]
    \centering
    {Subject-verb agreement}
    \includegraphics[width=0.98\linewidth]{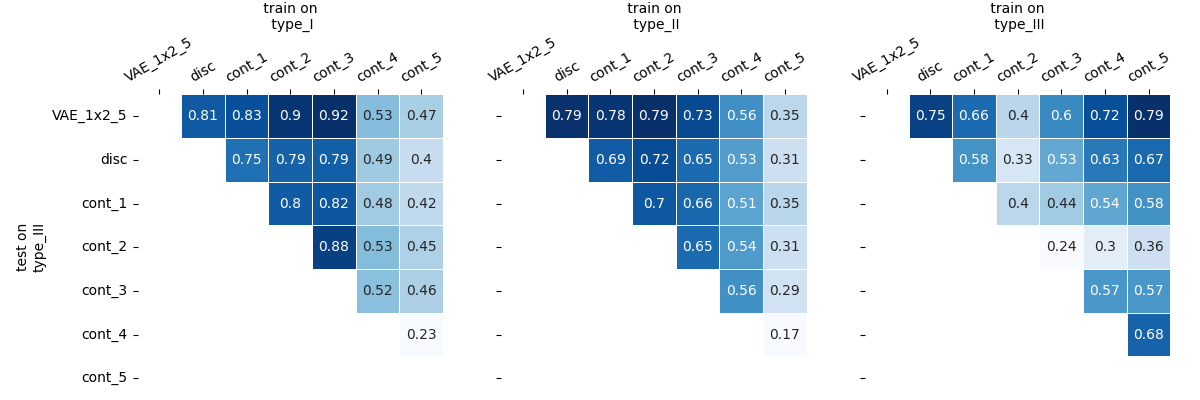}    
    {Verb alternations ALT-ATL}
    \includegraphics[width=0.98\linewidth]{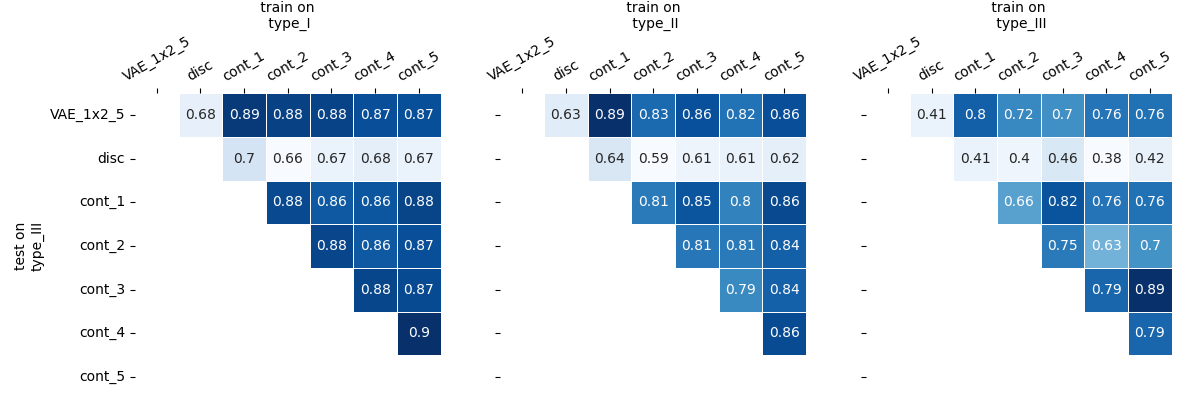}
    \caption{Errors when masking the latent vector in the joint sampling 1x2\_5 setting using Electra.}
    \label{fig:masked_agreement}
\end{figure}

%values ≤ 0 as indicating no agreement and 0.01–0.20 as none to slight, 0.21–0.40 as fair, 0.41– 0.60 as moderate, 0.61–0.80 as substantial, and 0.81–1.00 as almost perfect agreement

The lower the agreement the more different information the two settings encode. The heatmaps indicate that for the verb alternation problem, the discrete part of the latent encodes information that is most different from the base setup and all the continuous units. The distinction grows with lexical variation in the data -- it is highest when training on type III data. Masking the discrete part leads to a big increase in SSM errors (the syntax-semantic mapping), as shown in Figure \ref{fig:masked error} in the Appendix, which plots the absolute errors for the masked system variations. 

For the subject-verb agreement data, the continuous units encode the most distinct information, and this also becomes more pronounced with the increase in lexical variation. The error analysis in Figure \ref{fig:masked error} shows a high increase in WN2 errors when masking the units in the continuous part. This indicates a loss in long distance view of the model.

%The values corresponding to the discrete variable in the latent layer are constant after sampling. During masking, we change their values to their opposites ([1, 0] becomes [0, 1]) and decode accordingly. 

%The results show that not only the number of errors increases, but while the normal model makes specific errors, in the error patterns after masking the discrete part, \textcolor{red}{I don't see this}  \textit{the errors become more uniformly distributed. This indicates that the discrete part helps focus the system, and when errors are made, they are the ones most difficult to distinguish from the correct grammatical phenomenon. The continuous part of the latent layer encodes most of the relevant information. }

\paragraph{Discussion}
The performance on the multiple-choice problems and the error analysis show that including a discrete part for the latent layer in an encoder-decoder architecture leads to better results. The error analyses indicate that important information is captured in the discrete and continuous sections of the latent layer. Depending on the problem, either the discrete or the continuous latent sections contain more distinct information from each other. This shows that linguistic signals could be separated with such an architecture. We plan future work to enforce stronger disentanglement of the signals from the sentence embeddings, that can be linked to specific symbolic information.

\section{Conclusion}

Sentence embeddings combine a multitude of semantic and syntactic information in a continuous vector. We presented work that aims to disentangle such different linguistic signals from the sentence representation. We used diagnostic datasets, that focus on specific phenomena and encode them in a variety of contexts. The phenomena to discover are not explicitly provided, but are implied by the correct answer to a problem instance. We combined this data with a VAE-based system, and showed that we can induce a representation on the latent layer  that captures linguistic signals relevant to the targeted phenomena. Error analysis shows that the different parts of the latent layer captures slightly different signals. 

The consistent results of the same experimental set-up on different transformer-based sentence embeddings, on two different linguistic phenomena in two different languages supports our hypothesis that linguistic information is regularly distributed in the sentence embedding, and is retrievable and possibly ultimately mappable onto a more symbolic representation. We plan future work that forces more disentanglement of the signals encoded in the latent layer of the VAE-based system.

%We hypothesised that sentence embeddings produced by transformer-based models can be reinterpreted to separate specific grammatical information from the rest of the information encoded in the distributed representation. Experiments using systems based on the variational autoencoder architecture and datasets specifically built that encode different types of grammatical information -- subject-verb agreement, and verb alternations -- show that separating discrete and continuous signals on the latent layer lead to better detection of the targeted phenomena. In future work, we plan to disentangle the continuous portion of the latent layer. The systems were not specifically tuned for each type of grammatical information, but the results are consistent for both the targeted grammatical phenomena. This further supports our hypothesis, and indicates that grammatical information is regularly distributed in the sentence embedding, and is retrievable and possibly ultimately  mappable onto a more symbolic representation.

\section*{Limitations}

We performed experiments on an artificially generated dataset, that presents a grammatical phenomenon in a particular way -- as a sequence of sentences with specific properties. In future work we plan to separate the distillation of rules from a sentence representation from the processing of the sequence. 

We adopted the hyperparameters of the tested systems from previous experiments using sentence embeddings from a pretrained BERT model. This was a deliberate choice, as our goal was to investigate general properties of sentence embeddings with respect to different grammatical phenomena through the same systems. Specifically optimizing each architecture for each problem may lead to better individual results.

\section*{Ethics Statement}
To the best of our knowledge, there are no ethics concerns with this paper. 

%\section*{Acknowledgements}

\bibliography{anthology,custom}
\bibliographystyle{acl_natbib}

\newpage
\appendix

\onecolumn
\section{Supplementary Materials}

\subsection{Detailed results}

\begin{figure}[h!]
    \centering
    {Subject-verb agreement}
    
    \includegraphics[trim={0 1.2cm 0 0},clip,width=0.6\textwidth,height=6cm]{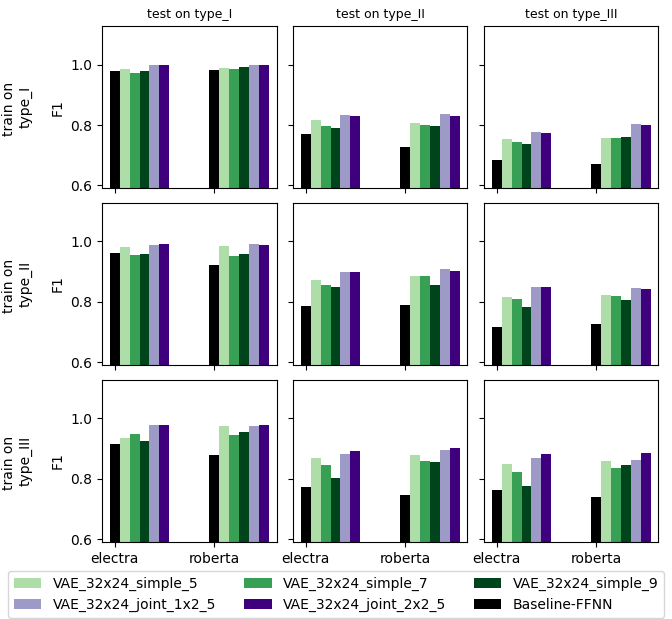} 
    
    {Verb alternations ALT-ATL}
    
    \includegraphics[trim={0 1.2cm 0 0},clip,width=0.6\textwidth,height=6cm]{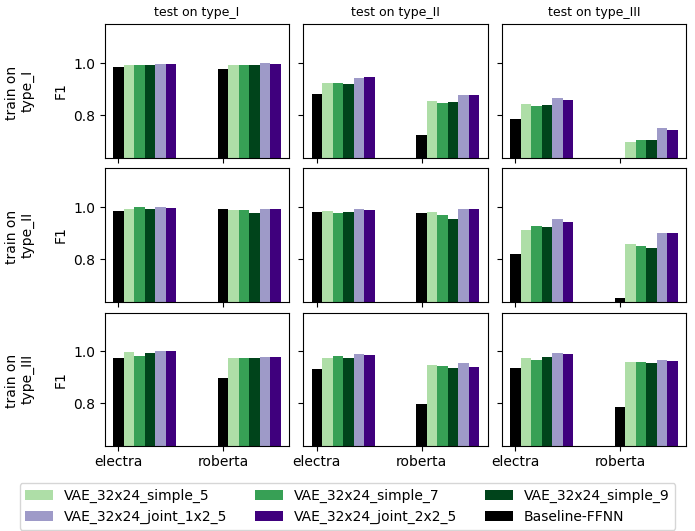}    
    
    {Verb alternations ATL-ALT}
    
    \includegraphics[width=0.6\textwidth,height=7cm]{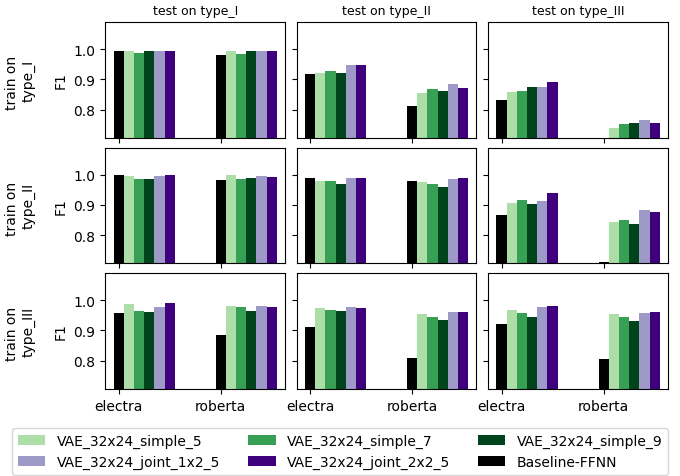}    
    \caption{F1 (avg. over 5 runs): continuous and joint samp., Electra and RoBERTa sentence embeddings.}
    \label{fig:agr_sampling-g2}
\end{figure}

Figure \ref{fig:agr_sampling-g2} shows the complete results, in terms of F1 averages over 5 runs (the standard deviation is less than 1e-03, so we do not include it), for all settings considered, and Electra and RoBERTa sentence embeddings.

\begin{figure*}[h!]
    \centering
    {Subject-verb agreement}
    \includegraphics[trim={0 1.2cm 0 0},clip,width=0.95\textwidth,height=0.26\textheight]{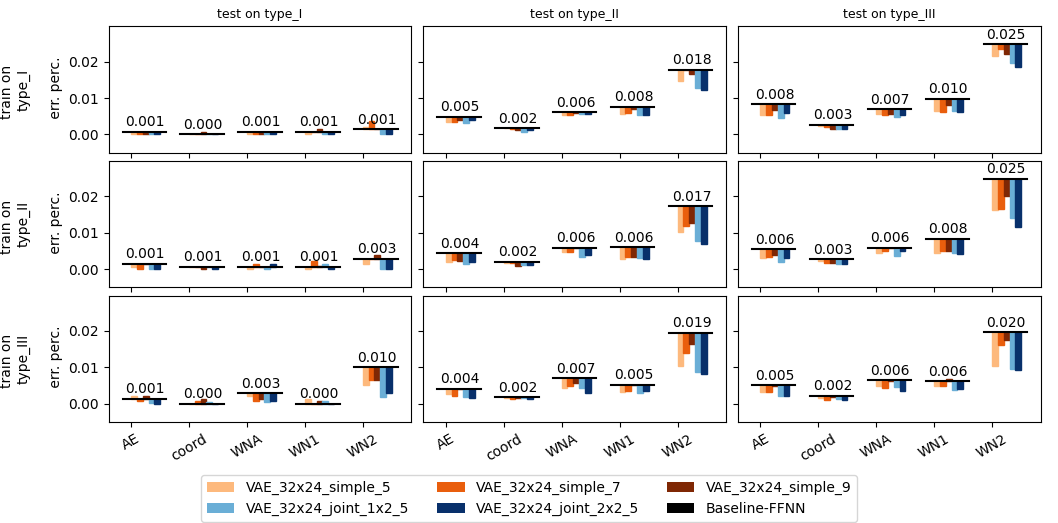}    
    {Verb alternations ALT-ATL}
    \includegraphics[trim={0 1.2cm 0 0},clip,width=0.95\textwidth,height=0.26\textheight]{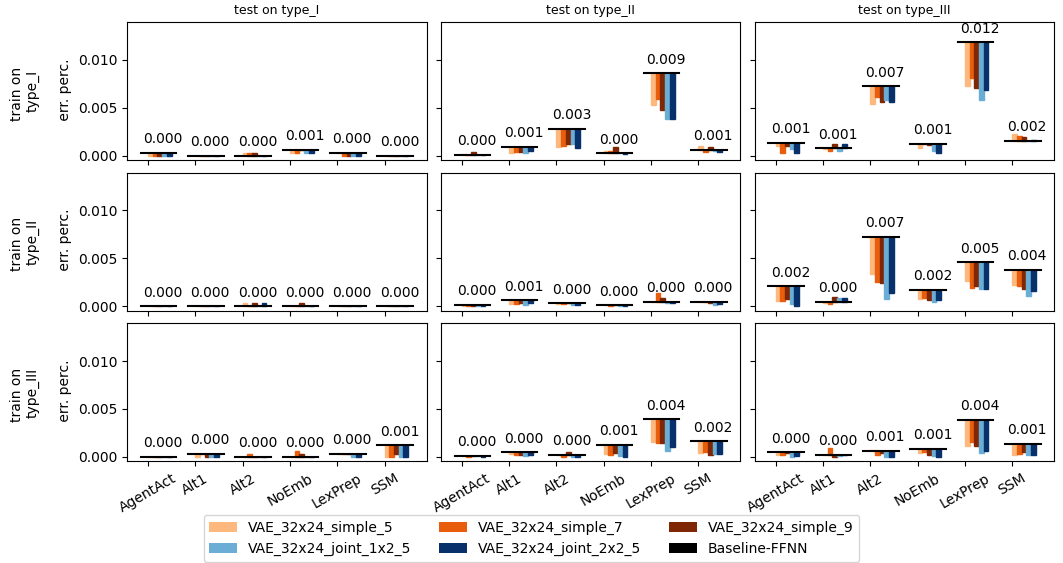}  
    \textbf{Verb alternations ATL-ALT}
    \includegraphics[width=0.95\textwidth,height=0.29\textheight]{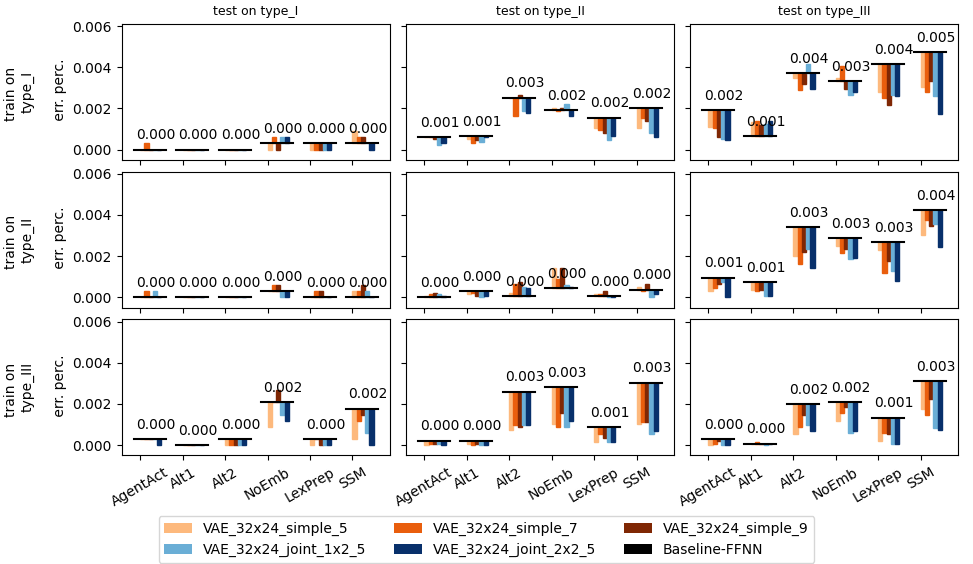}  
    \caption{Error analysis for continuous and joint sampling using Electra sentence embeddings.}
    \label{fig:errors-g2}
\end{figure*}

~\\
Figure \ref{fig:errors-g2} shows the error percentages for all settings, and both types of sentence embeddings.

\begin{figure*}[h!]
    \centering
    {Subject-verb agreement}
    \includegraphics[trim={0 1cm 0 0},clip,width=0.98\textwidth]{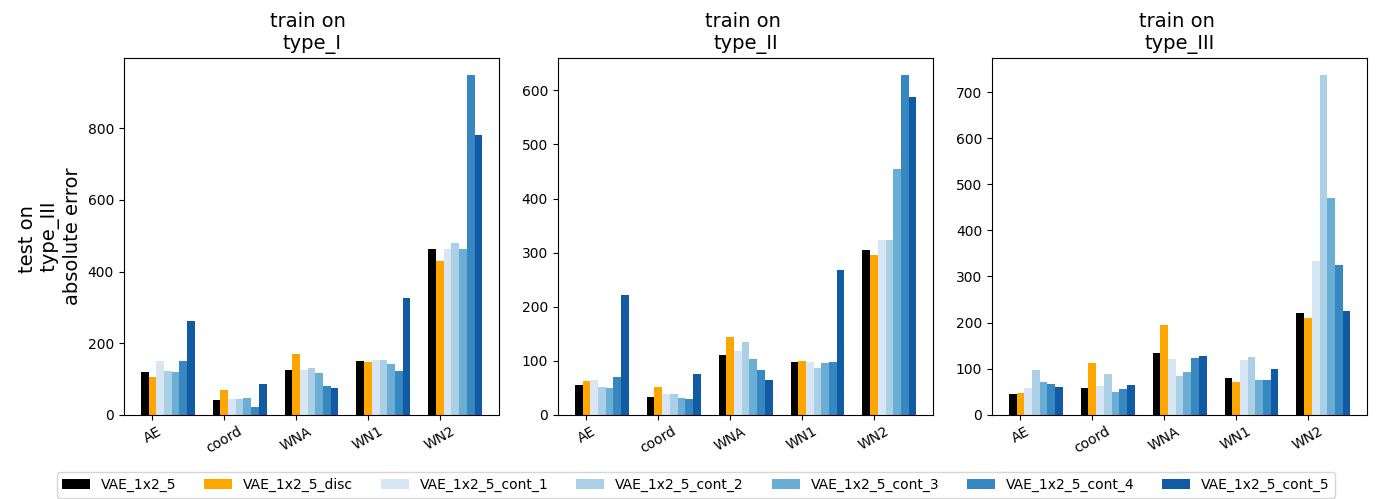}    
    {Verb alternations ALT-ATL}
    \includegraphics[trim={0 0.8cm 0 0},clip,width=0.98\textwidth]{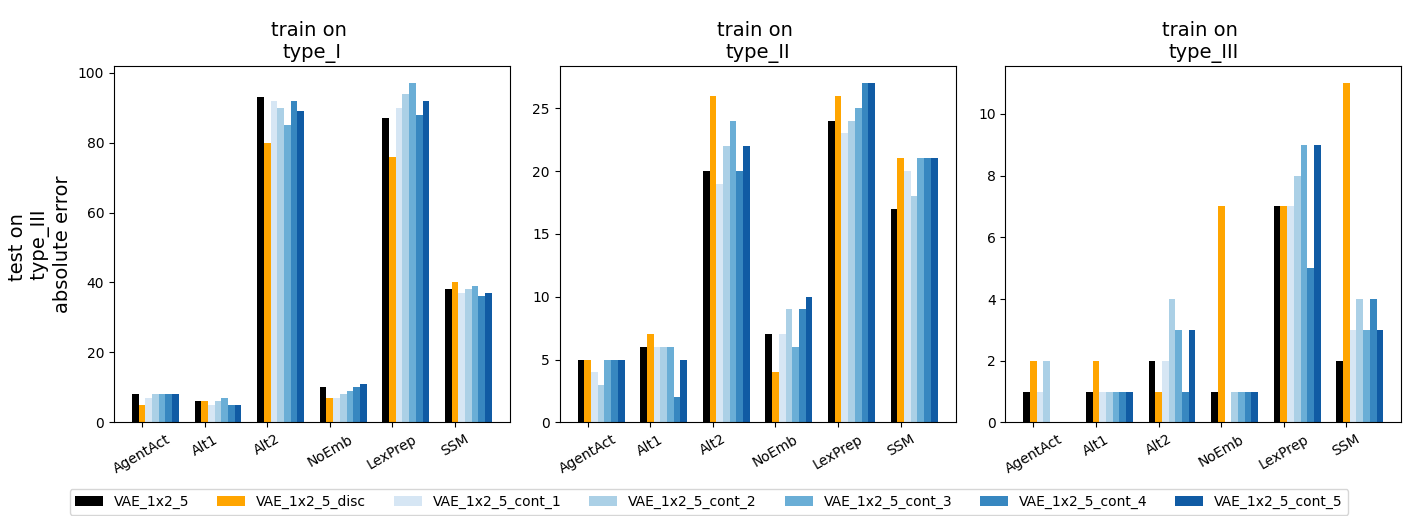}
    {Verb alternations ATL-ALT}
    \includegraphics[trim={0 0 0 0},clip,width=0.98\textwidth]{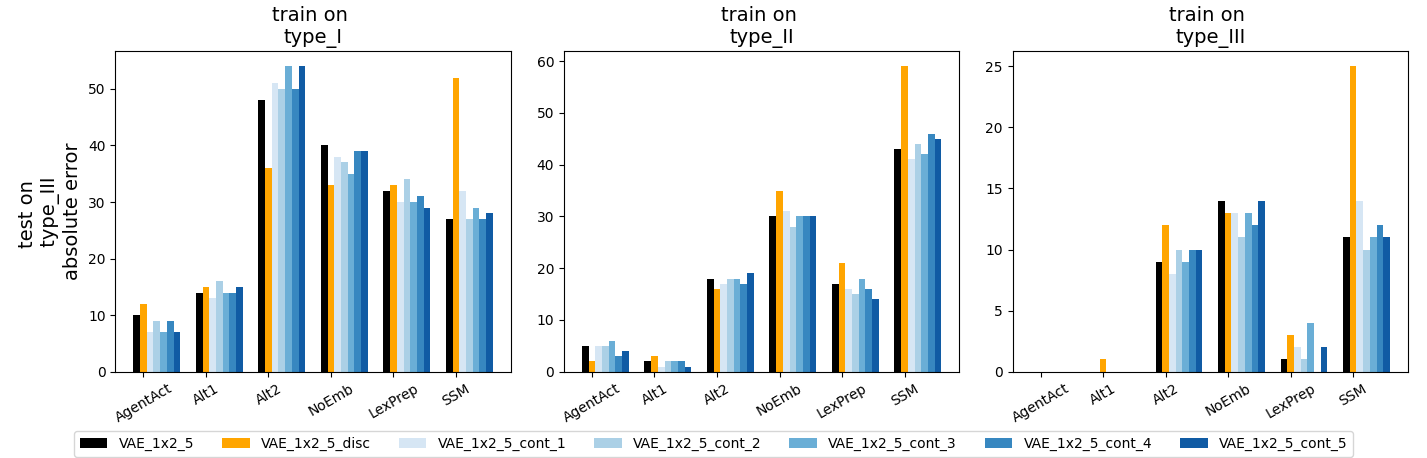}
    \caption{Errors when masking the latent vector in the joint sampling 1x2\_5 setting using Electra.}
    \label{fig:masked error}
\end{figure*}

~\\
Figure \ref{fig:masked error} shows the plot of absolute errors for a base system -- encoder decoder with joint sampling -- 1x2 (one binary category) + 5 (continuous units). The discrete part and each continuous unit are separately masked (set to 0), and the test data is then used to generate predictions. The plots shows the errors for the base system (black), and each masked variation.

\newpage

\begin{figure*}[h!]
    \centering
    {Subject-verb agreement}
    \includegraphics[width=0.98\textwidth]{figs/AGR_electra_masking_agreement.png}    
    {Verb alternations ALT-ATL}
    \includegraphics[width=0.98\textwidth]{figs/VB_ALT_gr1_electra_masking_agreement.png}
    {Verb alternations ATL-ALT}
    \includegraphics[width=0.98\textwidth]{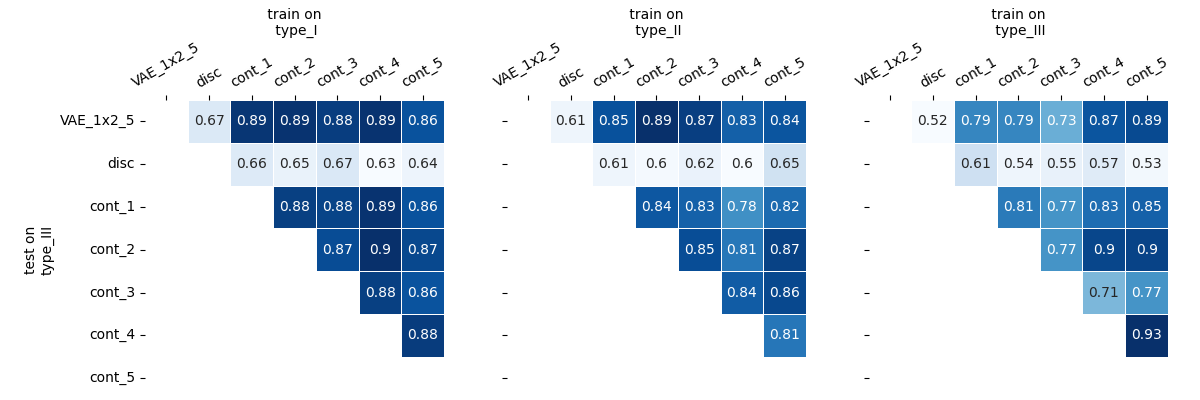}
    \caption{Errors when masking the latent vector in the joint sampling 1x2\_5 setting using Electra.}
    \label{fig:masked_agreement_all}
\end{figure*}

Figure \ref{fig:masked_agreement_all} shows the inter annotator agreement in terms of Cohen's $\kappa$, when the base system and each masked variation is considered an annotator. 

\end{document}

%% file: BLM-example.tex
  \begin{figure}[h]
  \small
  \setlength{\tabcolsep}{1mm}
    \begin{tabular}{lllll} 
    \hline
    \multicolumn{5}{c}{\sc Example of  context}\\
    	\hline
    	1 & The vase  & with the flower  &  & leaks. \\
    	2 & The vases & with the flower  &  & leak.\\
    	3 & The vase  & with the flowers &  & leaks. \\
    	4 & The vases & with the flowers &  & leak.\\
    	5 & The vase  & with the flower  & from the garden & leaks. \\
    	6 & The vases & with the flower  & from the garden & leak.\\
    	7 & The vase  & with the flowers & from the garden & leaks. \\
    	8 & ???
    \end{tabular}

  \setlength{\tabcolsep}{1mm}
    \begin{tabularx}{0.48\textwidth}{Xl} 
    \hline
    \multicolumn{2}{c}{\sc Example of  answers}\\
    	\hline
    The vase with the flower  and the garden  leaks. & Coord\\
    \textbf{The vases with the flowers  from the garden  leak.} & Correct\\
    The vase with the flower   leaks. & WNA \\
    The vase with the flower   from the garden  leak. & AE\\
    The vases with the flower  from the garden  leak. & WN1\\
    The vases with the flowers  from the gardens  leak. & WN2\\
    \hline 
    \end{tabularx}
    \caption{Subject-verb agreement BLM: a type I data instance (original in French). 
WNA=wrong nr. of attractors; AE=agreement error; WN1=wrong nr. for 1$^{st}$ attractor (N1); WN2=wrong nr. for 2$^{nd}$ attractor (N2).
 % See more examples in the supplementary materials.
   }
    \label{BLM-agreement}
  \end{figure}

%% file: BLMsl-example.tex
  \begin{figure}[h]
  \small
  \setlength{\tabcolsep}{1mm}
    \begin{tabular}{ll} 
    \hline
    \multicolumn{2}{c}{\sc Example of  context}\\
    	\hline
    	1 & The girl sprayed the wall with paint. \\
        2 & Paint was sprayed by the girl \\
        3 & Paint was sprayed onto the wall by the girl \\
        4 & Paint was sprayed onto the wall \\
        5 & The wall was sprayed by the girl \\
        6 & The wall was sprayed with the paint by the girl \\
        7 & The wall was sprayed with paint \\
    	8 & ???
    \end{tabular}
    
  \setlength{\tabcolsep}{1mm}
    \begin{tabular}{ll} 
    \hline
    \multicolumn{2}{c}{\sc Example of  answers}\\
    	\hline
    The girl sprayed paint onto the wall & Correct\\
    The girl was sprayed paint onto the wall & AgentAct \\
    The girl sprayed paint the wall & Alt1 \\
    The girl sprayed with paint onto the wall & Alt2 \\
    The girl sprayed paint for the room & NoEmb \\
    The girl sprayed paint under the wall & LexPrep\\
    Paint sprayed the girl onto the wall & SSM \\
    The wall sprayed the girl with paint & SSM \\
    Paint sprayed the wall with the girl & AASSM\\
%    The wall sprayed paint with the girl & AASSM\\ 
    \hline 
    \end{tabular}
    %  \end{minipage}\hfill
%  \begin{minipage}[r]{0.3\textwidth}
 \caption{Verb alternations. The labels indicate which (sub)rules are corrupted to create the error:
 % See more examples in the supplementary materials. 
 AgentAct=The agent in the alternant should be an NP in an active sentence; Alt=the alternation consists of a NP and a PP after the verb; NoEmb=the PP should not be embedded in the PP; LexPrep=the argument structure require given prepositions; SSM=syntax/semantic mapping; AASSM=simultaneous violations of Agent Act and SSM.}
    \label{BLM-sprayload}
%  \end{minipage}
  %  \end{center}
%    \vspace{-0.5cm}
\end{figure}

%% file: main.bbl
\begin{thebibliography}{41}
\expandafter\ifx\csname natexlab\endcsname\relax\def\natexlab#1{#1}\fi

\bibitem[{An et~al.(2023)An, Jiang, A.~Rodriguez, Nastase, and
  Merlo}]{an-etal-2023-blm}
Aixiu An, Chunyang Jiang, Maria A.~Rodriguez, Vivi Nastase, and Paola Merlo.
  2023.
\newblock \href {https://aclanthology.org/2023.eacl-main.99} {{BLM}-{A}gr{F}: A
  new {F}rench benchmark to investigate generalization of agreement in neural
  networks}.
\newblock In \emph{Proceedings of the 17th Conference of the European Chapter
  of the Association for Computational Linguistics}, pages 1363--1374,
  Dubrovnik, Croatia. Association for Computational Linguistics.

\bibitem[{Bao et~al.(2019)Bao, Zhou, Huang, Li, Mou, Vechtomova, Dai, and
  Chen}]{bao-etal-2019-generating}
Yu~Bao, Hao Zhou, Shujian Huang, Lei Li, Lili Mou, Olga Vechtomova, Xin-yu Dai,
  and Jiajun Chen. 2019.
\newblock \href {https://doi.org/10.18653/v1/P19-1602} {Generating sentences
  from disentangled syntactic and semantic spaces}.
\newblock In \emph{Proceedings of the 57th Annual Meeting of the Association
  for Computational Linguistics}, pages 6008--6019, Florence, Italy.
  Association for Computational Linguistics.

\bibitem[{Bengio et~al.(2013)Bengio, Courville, and Vincent}]{bengio-ea2013}
Yoshua Bengio, Aaron Courville, and Pascal Vincent. 2013.
\newblock Representation learning: A review and new perspectives.
\newblock \emph{IEEE transactions on pattern analysis and machine
  intelligence}, 35(8):1798–1828.

\bibitem[{Chen et~al.(2019{\natexlab{a}})Chen, Tang, Wiseman, and
  Gimpel}]{chen-etal-2019-multi}
Mingda Chen, Qingming Tang, Sam Wiseman, and Kevin Gimpel. 2019{\natexlab{a}}.
\newblock \href {https://doi.org/10.18653/v1/N19-1254} {A multi-task approach
  for disentangling syntax and semantics in sentence representations}.
\newblock In \emph{Proceedings of the 2019 Conference of the North {A}merican
  Chapter of the Association for Computational Linguistics: Human Language
  Technologies, Volume 1 (Long and Short Papers)}, pages 2453--2464,
  Minneapolis, Minnesota. Association for Computational Linguistics.

\bibitem[{Chen et~al.(2019{\natexlab{b}})Chen, Li, Grosse, and
  Duvenaud}]{chen2615isolating}
Ricky~TQ Chen, Xuechen Li, Roger Grosse, and David Duvenaud.
  2019{\natexlab{b}}.
\newblock Isolating sources of disentanglement in vaes.
\newblock In \emph{Proceedings of the 32nd International Conference on Neural
  Information Processing Systems}.

\bibitem[{Clark et~al.(2020)Clark, Luong, Le, and Manning}]{clark2020electra}
Kevin Clark, Minh-Thang Luong, Quoc~V. Le, and Christopher~D. Manning. 2020.
\newblock \href {https://openreview.net/pdf?id=r1xMH1BtvB} {{ELECTRA}:
  Pre-training text encoders as discriminators rather than generators}.
\newblock In \emph{ICLR}.

\bibitem[{Devlin et~al.(2019)Devlin, Chang, Lee, and
  Toutanova}]{devlin-etal-2019-bert}
Jacob Devlin, Ming-Wei Chang, Kenton Lee, and Kristina Toutanova. 2019.
\newblock \href {https://doi.org/10.18653/v1/N19-1423} {{BERT}: Pre-training of
  deep bidirectional transformers for language understanding}.
\newblock In \emph{Proceedings of the 2019 Conference of the North {A}merican
  Chapter of the Association for Computational Linguistics: Human Language
  Technologies, Volume 1 (Long and Short Papers)}, pages 4171--4186,
  Minneapolis, Minnesota. Association for Computational Linguistics.

\bibitem[{Dupont(2018)}]{dupont2018learning}
Emilien Dupont. 2018.
\newblock Learning disentangled joint continuous and discrete representations.
\newblock \emph{Advances in Neural Information Processing Systems}, 31.

\bibitem[{Fu et~al.(2018)Fu, Tan, Peng, Zhao, and Yan}]{Fu2018style-transfer}
Zhenxin Fu, Xiaoye Tan, Nanyun Peng, Dongyan Zhao, and Rui Yan. 2018.
\newblock \href {https://doi.org/10.1609/aaai.v32i1.11330} {Style transfer in
  text: Exploration and evaluation}.
\newblock \emph{Proceedings of the AAAI Conference on Artificial Intelligence},
  32(1).

\bibitem[{Goldberg(2019)}]{goldberg2019}
Yoav Goldberg. 2019.
\newblock Assessing bert's syntactic abilities.
\newblock \emph{arXiv preprint arXiv:1901.05287}.

\bibitem[{Goodfellow et~al.(2014)Goodfellow, Pouget-Abadie, Mirza, Xu,
  Warde-Farley, Ozair, Courville, and Bengio}]{goodfellowNIPS2014}
Ian Goodfellow, Jean Pouget-Abadie, Mehdi Mirza, Bing Xu, David Warde-Farley,
  Sherjil Ozair, Aaron Courville, and Yoshua Bengio. 2014.
\newblock \href
  {https://proceedings.neurips.cc/paper_files/paper/2014/file/5ca3e9b122f61f8f06494c97b1afccf3-Paper.pdf}
  {Generative adversarial nets}.
\newblock In \emph{Advances in Neural Information Processing Systems},
  volume~27. Curran Associates, Inc.

\bibitem[{Gulordava et~al.(2018)Gulordava, Bojanowski, Grave, Linzen, and
  Baroni}]{gulordava2018}
Kristina Gulordava, Piotr Bojanowski, Edouard Grave, Tal Linzen, and Marco
  Baroni. 2018.
\newblock \href {https://doi.org/10.18653/v1/N18-1108} {Colorless green
  recurrent networks dream hierarchically}.
\newblock In \emph{Proceedings of the 2018 Conference of the North American
  Chapter of the Association for Computational Linguistics: Human Language
  Technologies}, pages 1195--1205. Association for Computational Linguistics.

\bibitem[{Higgins et~al.(2017)Higgins, Matthey, Pal, Burgess, Glorot,
  Botvinick, Mohamed, and Lerchner}]{higgins2017beta}
Irina Higgins, Loic Matthey, Arka Pal, Christopher Burgess, Xavier Glorot,
  Matthew Botvinick, Shakir Mohamed, and Alexander Lerchner. 2017.
\newblock beta-vae: Learning basic visual concepts with a constrained
  variational framework.
\newblock In \emph{International Conference on Learning Representations
  (ICLR)}.

\bibitem[{Huang et~al.(2021)Huang, Huang, and
  Chang}]{huang-etal-2021-disentangling}
James~Y. Huang, Kuan-Hao Huang, and Kai-Wei Chang. 2021.
\newblock \href {https://doi.org/10.18653/v1/2021.naacl-main.108}
  {Disentangling semantics and syntax in sentence embeddings with pre-trained
  language models}.
\newblock In \emph{Proceedings of the 2021 Conference of the North American
  Chapter of the Association for Computational Linguistics: Human Language
  Technologies}, pages 1372--1379, Online. Association for Computational
  Linguistics.

\bibitem[{Jang et~al.(2017)Jang, Gu, and Poole}]{jang2017categorical}
Eric Jang, Shixiang Gu, and Ben Poole. 2017.
\newblock Categorical reparameterization with gumbel-softmax.

\bibitem[{John et~al.(2019)John, Mou, Bahuleyan, and
  Vechtomova}]{john-etal-2019-disentangled}
Vineet John, Lili Mou, Hareesh Bahuleyan, and Olga Vechtomova. 2019.
\newblock \href {https://doi.org/10.18653/v1/P19-1041} {Disentangled
  representation learning for non-parallel text style transfer}.
\newblock In \emph{Proceedings of the 57th Annual Meeting of the Association
  for Computational Linguistics}, pages 424--434, Florence, Italy. Association
  for Computational Linguistics.

\bibitem[{Kann et~al.(2019)Kann, Warstadt, Williams, and
  Bowman}]{kann-etal-2019-verb}
Katharina Kann, Alex Warstadt, Adina Williams, and Samuel~R. Bowman. 2019.
\newblock \href {https://doi.org/10.7275/q5js-4y86} {Verb argument structure
  alternations in word and sentence embeddings}.
\newblock In \emph{Proceedings of the Society for Computation in Linguistics
  ({SC}i{L}) 2019}, pages 287--297.

\bibitem[{Kingma et~al.(2015)Kingma, Salimans, and Welling}]{kingma2015}
Diederik~P Kingma, Tim Salimans, and Max Welling. 2015.
\newblock \href
  {https://proceedings.neurips.cc/paper/2015/file/bc7316929fe1545bf0b98d114ee3ecb8-Paper.pdf}
  {Variational dropout and the local reparameterization trick}.
\newblock In \emph{Advances in Neural Information Processing Systems},
  volume~28. Curran Associates, Inc.

\bibitem[{Kingma and Welling(2013)}]{kingma2013vae}
Diederik~P Kingma and Max Welling. 2013.
\newblock Auto-encoding variational bayes.
\newblock \emph{arXiv preprint arXiv:1312.6114}.

\bibitem[{Levin(1993)}]{levin1993}
Beth Levin. 1993.
\newblock \emph{{English} verb classes and alternations: A preliminary
  investigation}.
\newblock University of Chicago Press.

\bibitem[{Linzen and Baroni(2021)}]{linzen2021}
Tal Linzen and Marco Baroni. 2021.
\newblock \href {https://doi.org/10.1146/annurev-linguistics-032020-051035}
  {Syntactic structure from deep learning}.
\newblock \emph{Annual Review of Linguistics}, 7(1):195--212.

\bibitem[{Linzen et~al.(2016)Linzen, Dupoux, and Goldberg}]{linzen2016}
Tal Linzen, Emmanuel Dupoux, and Yoav Goldberg. 2016.
\newblock \href {https://www.mitpressjournals.org/doi/abs/10.1162/tacl_a_00115}
  {Assessing the ability of {LSTMs} to learn syntax-sensitive dependencies}.
\newblock \emph{Transactions of the Association of Computational Linguistics},
  4(1):521--535.

\bibitem[{Liu et~al.(2019)Liu, Ott, Goyal, Du, Joshi, Chen, Levy, Lewis,
  Zettlemoyer, and Stoyanov}]{liu2019roberta}
Yinhan Liu, Myle Ott, Naman Goyal, Jingfei Du, Mandar Joshi, Danqi Chen, Omer
  Levy, Mike Lewis, Luke Zettlemoyer, and Veselin Stoyanov. 2019.
\newblock Roberta: A robustly optimized bert pretraining approach.
\newblock \emph{arXiv preprint arXiv:1907.11692}.

\bibitem[{Locatello et~al.(2020)Locatello, Poole, Raetsch, Sch{\"o}lkopf,
  Bachem, and Tschannen}]{pmlr-v119-locatello20a}
Francesco Locatello, Ben Poole, Gunnar Raetsch, Bernhard Sch{\"o}lkopf, Olivier
  Bachem, and Michael Tschannen. 2020.
\newblock \href {https://proceedings.mlr.press/v119/locatello20a.html}
  {Weakly-supervised disentanglement without compromises}.
\newblock In \emph{Proceedings of the 37th International Conference on Machine
  Learning}, volume 119 of \emph{Proceedings of Machine Learning Research},
  pages 6348--6359. PMLR.

\bibitem[{Mathieu et~al.(2019)Mathieu, Rainforth, Siddharth, and
  Teh}]{mathieu2019disentangling}
Emile Mathieu, Tom Rainforth, Nana Siddharth, and Yee~Whye Teh. 2019.
\newblock Disentangling disentanglement in variational autoencoders.
\newblock In \emph{International Conference on Machine Learning}, pages
  4402--4412. PMLR.

\bibitem[{Mercatali and
  Freitas(2021)}]{mercatali-freitas-2021-disentangling-generative}
Giangiacomo Mercatali and Andr{\'e} Freitas. 2021.
\newblock \href {https://doi.org/10.18653/v1/2021.findings-emnlp.301}
  {Disentangling generative factors in natural language with discrete
  variational autoencoders}.
\newblock In \emph{Findings of the Association for Computational Linguistics:
  EMNLP 2021}, pages 3547--3556, Punta Cana, Dominican Republic. Association
  for Computational Linguistics.

\bibitem[{Merlo(2023)}]{merlo2023}
Paola Merlo. 2023.
\newblock \href {https://doi.org/10.48550/arXiv.2306.11444} {Blackbird language
  matrices {(BLM)}, a new task for rule-like generalization in neural networks:
  Motivations and formal specifications}.
\newblock \emph{ArXiv}, cs.CL 2306.11444.

\bibitem[{Merlo et~al.(2022)Merlo, An, and Rodriguez}]{merlo-ea2022arxiv}
Paola Merlo, Aixiu An, and Maria~A. Rodriguez. 2022.
\newblock \href {https://doi.org/10.48550/ARXIV.2205.10866} {Blackbird's
  language matrices {(BLMs)}: a new benchmark to investigate disentangled
  generalisation in neural networks}.
\newblock \emph{ArXiv}, cs.CL 2205.10866.

\bibitem[{Nastase and Merlo(2023)}]{nastase-merlo2023}
Vivi Nastase and Paola Merlo. 2023.
\newblock Grammatical information in {BERT} sentence embeddings as
  two-dimensional arrays.
\newblock In \emph{Proceedings of the 8th Workshop on Representation Learning
  for NLP (RepL4NLP 2023)}, Toronto, Canada.

\bibitem[{Nikolaev and Pad{\'o}(2023)}]{nikolaev-pado-2023-representation}
Dmitry Nikolaev and Sebastian Pad{\'o}. 2023.
\newblock \href {https://aclanthology.org/2023.eacl-main.268} {Representation
  biases in sentence transformers}.
\newblock In \emph{Proceedings of the 17th Conference of the European Chapter
  of the Association for Computational Linguistics}, pages 3701--3716,
  Dubrovnik, Croatia. Association for Computational Linguistics.

\bibitem[{Rajpurkar et~al.(2018)Rajpurkar, Jia, and
  Liang}]{rajpurkar-etal-2018-know}
Pranav Rajpurkar, Robin Jia, and Percy Liang. 2018.
\newblock \href {https://doi.org/10.18653/v1/P18-2124} {Know what you don{'}t
  know: Unanswerable questions for {SQ}u{AD}}.
\newblock In \emph{Proceedings of the 56th Annual Meeting of the Association
  for Computational Linguistics (Volume 2: Short Papers)}, pages 784--789,
  Melbourne, Australia. Association for Computational Linguistics.

\bibitem[{Rappaport and Levin(1988)}]{rappaport1988theta}
Malka Rappaport and Beth Levin. 1988.
\newblock What to do with theta-roles.
\newblock In Wendy Wilkins, editor, \emph{Thematic relations}, pages 7--36.

\bibitem[{Raven(1938)}]{raven1938}
John~C. Raven. 1938.
\newblock Standardization of progressive matrices.
\newblock \emph{British Journal of Medical Psychology}, 19:137--150.

\bibitem[{Rudin et~al.(2021)Rudin, Chen, Chen, Huang, Semenova, and
  Zhong}]{Rudin2021InterpretableML}
Cynthia Rudin, Chaofan Chen, Zhi Chen, Haiyang Huang, Lesia Semenova, and Chudi
  Zhong. 2021.
\newblock Interpretable machine learning: Fundamental principles and 10 grand
  challenges.
\newblock \emph{ArXiv}, abs/2103.11251.

\bibitem[{Samo et~al.(2023)Samo, Nastase, Jiang, and Merlo}]{samo-etal-2023}
Giuseppe Samo, Vivi Nastase, Chunyang Jiang, and Paola Merlo. 2023.
\newblock {BLM-s/lE}: A structured dataset of {English} spray-load verb
  alternations for testing generalization in {LLM}s.
\newblock In \emph{Findings of the 2023 Conference on Empirical Methods in
  Natural Language Processing}.

\bibitem[{Schmidhuber(1992)}]{schmidhuber1992}
Jürgen Schmidhuber. 1992.
\newblock \href {https://doi.org/10.1162/neco.1992.4.6.863} {{Learning
  Factorial Codes by Predictability Minimization}}.
\newblock \emph{Neural Computation}, 4(6):863--879.

\bibitem[{Wang et~al.(2019)Wang, Pruksachatkun, Nangia, Singh, Michael, Hill,
  Levy, and Bowman}]{wang-ea2019}
Alex Wang, Yada Pruksachatkun, Nikita Nangia, Amanpreet Singh, Julian Michael,
  Felix Hill, Omer Levy, and Samuel Bowman. 2019.
\newblock \href
  {https://proceedings.neurips.cc/paper/2019/file/4496bf24afe7fab6f046bf4923da8de6-Paper.pdf}
  {Superglue: A stickier benchmark for general-purpose language understanding
  systems}.
\newblock In \emph{Advances in Neural Information Processing Systems},
  volume~32. Curran Associates, Inc.

\bibitem[{Yi et~al.(2022)Yi, Bruno, Han, Zukerman, and
  Steinert-Threlkeld}]{yi-etal-2022-probing}
David Yi, James Bruno, Jiayu Han, Peter Zukerman, and Shane Steinert-Threlkeld.
  2022.
\newblock \href {https://aclanthology.org/2022.blackboxnlp-1.12} {Probing for
  understanding of {E}nglish verb classes and alternations in large pre-trained
  language models}.
\newblock In \emph{Proceedings of the Fifth BlackboxNLP Workshop on Analyzing
  and Interpreting Neural Networks for NLP}, pages 142--152, Abu Dhabi, United
  Arab Emirates (Hybrid). Association for Computational Linguistics.

\bibitem[{Zhang et~al.(2019)Zhang, Gao, Jia, Zhu, and Zhu}]{chi2019RAVEN}
Chi Zhang, Feng Gao, Baoxiong Jia, Yixin Zhu, and Song-Chun Zhu. 2019.
\newblock Raven: A dataset for relational and analogical visual reasoning.
\newblock In \emph{Proceedings of the IEEE Conference on Computer Vision and
  Pattern Recognition (CVPR)}.

\bibitem[{Zheng and Lapata(2022)}]{zheng-lapata-2022-disentangled}
Hao Zheng and Mirella Lapata. 2022.
\newblock \href {https://doi.org/10.18653/v1/2022.acl-long.293} {Disentangled
  sequence to sequence learning for compositional generalization}.
\newblock In \emph{Proceedings of the 60th Annual Meeting of the Association
  for Computational Linguistics (Volume 1: Long Papers)}, pages 4256--4268,
  Dublin, Ireland. Association for Computational Linguistics.

\bibitem[{Zhou and Neubig(2017)}]{zhou-neubig-2017-multi}
Chunting Zhou and Graham Neubig. 2017.
\newblock \href {https://doi.org/10.18653/v1/P17-1029} {Multi-space variational
  encoder-decoders for semi-supervised labeled sequence transduction}.
\newblock In \emph{Proceedings of the 55th Annual Meeting of the Association
  for Computational Linguistics (Volume 1: Long Papers)}, pages 310--320,
  Vancouver, Canada. Association for Computational Linguistics.

\end{thebibliography}
